\documentclass[sigconf]{aamas}
\usepackage{balance} % for balancing columns on the final page
\usepackage{graphicx}
\usepackage{caption}
\usepackage{subcaption}
\usepackage{amsmath}
\usepackage{amsfonts}
\usepackage{multirow}
\usepackage{balance}
\usepackage{newfloat}
\usepackage{pbox}
\usepackage{algorithmic}
\usepackage{algorithm2e}
\usepackage{listings}
\setcopyright{none}
\renewcommand\footnotetextcopyrightpermission[1]{}
\title{Using General Value Functions to Learn Domain-Backed Inventory Management Policies}

\author{Durgesh Kalwar}
\affiliation{
  \institution{TCS Research}
  \city{Mumbai}
  \country{India}}
\email{durgesh.kalwar@tcs.com}

\author{Omkar Shelke}
\affiliation{
  \institution{TCS Research}
  \city{Mumbai}
  \country{India}}
\email{shelke.omkar@tcs.com}

\author{Harshad Khadilkar}
\affiliation{
  \institution{TCS Research}
  \city{Mumbai}
  \country{India}}
\email{harshad.khadilkar@tcs.com}

\begin{abstract}
We consider the inventory management problem, where the goal is to balance conflicting objectives such as availability and wastage of a large range of products in a store. We propose a reinforcement learning (RL) approach that utilises General Value Functions (GVFs) to derive domain-backed inventory replenishment policies. The inventory replenishment decisions are modelled as a sequential decision making problem, which is challenging due to uncertain demand and the existence of aggregate (cross-product) constraints. In existing literature, GVFs have primarily been used for auxiliary task learning. We use this capability to train GVFs on domain-critical characteristics such as prediction of stock-out probability and wastage quantity. Using this domain expertise for more effective exploration, we train an RL agent to compute the inventory replenishment quantities for a large range of products (up to 6000 in the reported experiments), which share aggregate constraints such as the total weight/volume per delivery. Additionally, we show that the GVF predictions can be used to provide additional domain-backed insights into the decisions proposed by the RL agent. Finally, since the environment dynamics are fully transferred, the trained GVFs can be used for faster adaptation to vastly different business objectives (for example, due to the start of a promotional period or due to deployment in a new customer environment).
\end{abstract}


\newcommand{\BibTeX}{\rm B\kern-.05em{\sc i\kern-.025em b}\kern-.08em\TeX}

\begin{document}

%%% The following commands remove the headers in your paper. For final 
%%% papers, these will be inserted during the pagination process.

\pagestyle{fancy}
\fancyhead{}

%%% The next command prints the information defined in the preamble.

\maketitle 

%%%%%%%%%%%%%%%%%%%%%%%%%%%%%%%%%%%%%%%%%%%%%%%%%%%%%%%%%%%%%%%%%%%%%%%%

\section{Introduction}

The Inventory Management (IM) problem is one of the most critical challenges within the supply chain industry \citep{nahmias1993mathematical}, particularly when it comes to replenishment decisions. Striking the right balance between maintaining adequate stock levels to meet the customer demands while minimizing inventory costs is a perpetual concern. Achieving this delicate equilibrium has profound implications for cost efficiency, customer satisfaction and overall supply chain performance. In the context of supply chain, the management of replenishment decisions is pivotal since it directly influences inventory turnover, lead times, and the ability to respond swiftly to fluctuations within demand. Efficient and data-driven strategies address this complex issue effectively and they continue to be a subject of rigorous research and practical innovation within the industry. Some of the formative work in this domain employing dynamic programming \citep{clark1960optimal}, provided a theoretical foundation for optimal base-stock policies in simple series systems. Nonetheless, the complexity of the recursive equations grows substantially with the problem's size, prompting the development of various approximation algorithms, such as stochastic approximation \citep{kushner2012stochastic}, infinitesimal perturbation analysis (IPA) \citep{proth1996marking}, and approximate dynamic programming \citep{bertsekas1996neuro}, to tackle this. However, the practical implementation of these methods remains a formidable challenge \citep{bertsimas2006robust}.

Traditional methods in inventory management (IM) have often simplified the problem by applying dynamic programming (DP) techniques. But, such approaches tend to make assumptions that may not hold in real-world scenarios. For instance, they often assume that customer demands are independently and identically distributed (iid) and that leading times are deterministic \citep{kaplan1970dynamic, ehrhardt1984s}. These assumptions, while convenient for mathematical modeling, can deviate significantly from actual conditions. Furthermore, as certain critical factors like leading times and the number of stock keeping units (SKUs) increase, the state space of the problem expands rapidly. This expansion, often referred to as the "curse of dimensionality" \citep{gijsbrechts2022can}, renders the problem intractable when addressed using DP. To overcome these limitations, various approaches based on approximate dynamic programming have been introduced to address IM challenges across diverse contexts \citep{halman2009fully, fang2013sourcing, chen2019heuristic}. While these alternative methods have demonstrated efficacy in specific scenarios, they tend to rely heavily on problem-specific expertise or make assumptions that might not be universally applicable. For instance, some of these approaches assume zero or one-period leading times \citep{halman2009fully}, which might not hold true in every inventory management setting. Consequently, their ability to generalize to broader IM contexts remains constrained.

Reinforcement Learning (RL) algorithms have proven effective in a wide array of applications involving sequential decision-making. The applications encompass areas such as software-based gaming \citep{doyle1988state, silver2017mastering}, as well as physical systems like autonomous driving \citep{shalev2016safe}, and flight dynamics \citep{ng2006autonomous}. A notable characteristic of these applications is their capacity to devise strategies that optimize long-term future rewards while adhering to constraints. This quality positions RL as a natural choice for a broader category of problems. We investigate its applicability in high-dimensional systems, particularly in real-world contexts like industrial operations. Our focus is on systems driven by vector differential equations, stemming from operations research. However, this approach can be adapted to any scenario requiring complex decision-making without online searching. RL methods are versatile for data-driven situations. Yet, creating a single policy for all stock keeping units (SKUs) is challenging due to the vast state and action space involved \citep{jiang2018open}. Two recent papers \citep{barat2019actor, sultana2020reinforcement} where more realistic scenarios containing multiple SKUs are considered. In \citep{barat2019actor}, the main contribution is to propose a framework to support efficient deployment of RL algorithms in real systems. As an example, the authors introduce a centralized algorithm for solving the IM problem. In contrast, a decentralized algorithm is proposed in \citep{sultana2020reinforcement} to solve the IM problem with multiple SKUs in a multi-echelon setting. These two papers use the off-the-shelf RL algorithm \citep{baselines} to solve for the policy. 

General Value Functions (GVFs) are crucial in reinforcement learning (RL), offering a versatile framework \citep{sutton2011horde} for estimating various value functions, from standard state-value and action-value functions to specialized RL objectives. GVFs empower agents to grasp the value of different signals or goals in an environment, enhancing their understanding of diverse aspects of the learning problem. They prove especially valuable for abstracting and generalizing information across tasks and scenarios, enabling efficient knowledge transfer, which is vital in complex, dynamic environments where a single value function may fall short. Recent RL advancements, like the Option-Critic Architecture \citep{bacon2017option} and successor representation \citep{kulkarni2016hierarchical}, rely on GVFs to boost learning efficiency across diverse tasks. In a recent study \citep{kalwar2022follow}, GVFs were applied in navigation environments with vast state spaces and sparse rewards. They were employed for learning auxiliary tasks as value functions and generating directed exploration actions.

In this paper, we aim to showcase the following contributions:

\begin{enumerate}
    \item We present an RL strategy which makes use of General Value Functions for learning the underlying dynamics of an inventory management problem, thus improving (i) the effectiveness of exploration during training, and (ii) the transferability of learnt embeddings and policies to other tasks within the same environment. Our approach is specifically trained on system attributes such as critical levels of inventory and wastage quantity generated.
    \item Experiments on scenarios ranging from 100 to 6000 products are carried out along with other baselines. We provide further insights into the GVF learnt models by portraying their predictions for different attributes. Also we showcase their generalization abilities by their quick assimilation to other domain objectives.   
\end{enumerate}
%%%%%%%%%%%%%%%%%%%%%%%%%%%%%%%%%%%%%%%%%%%%%%%%%%%%%%%%%%%%%%%%%%%%%%%%

% \section{The Preamble}

% You will be assigned a submission number when you register the abstract 
% of your paper on \textit{EasyChair}. Include this number in your 
% document using the `\verb|\acmSubmissionID|' command.

% Then use the familiar commands to specify the title and authors of your
% paper in the preamble of the document. The title should be appropriately 
% capitalised (meaning that every `important' word in the title should 
% start with a capital letter). For the final version of your paper, make 
% sure to specify the affiliation and email address of each author using 
% the appropriate commands. Specify an affiliation and email address 
% separately for each author, even if two authors share the same 
% affiliation. You can specify more than one affiliation for an author by 
% using a separate `\verb|\affiliation|' command for each affiliation.

% Provide a short abstract using the `\texttt{abstract}' environment.
 
% Finally, specify a small number of keywords characterising your work, 
% using the `\verb|\keywords|' command. 
\section{Problem Description}

\begin{figure}
    \centering
    \includegraphics[width=0.45\textwidth]{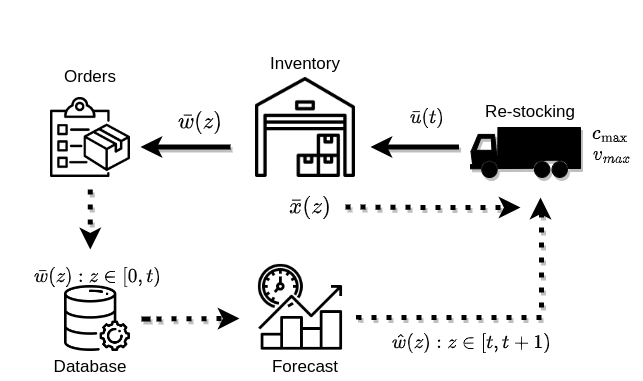}
    \caption{A simple flowchart of the inventory management problem at hand}
    \label{fig:im_problem}
\end{figure}

\begin{figure}
    \centering
    \includegraphics[width=0.49\textwidth]{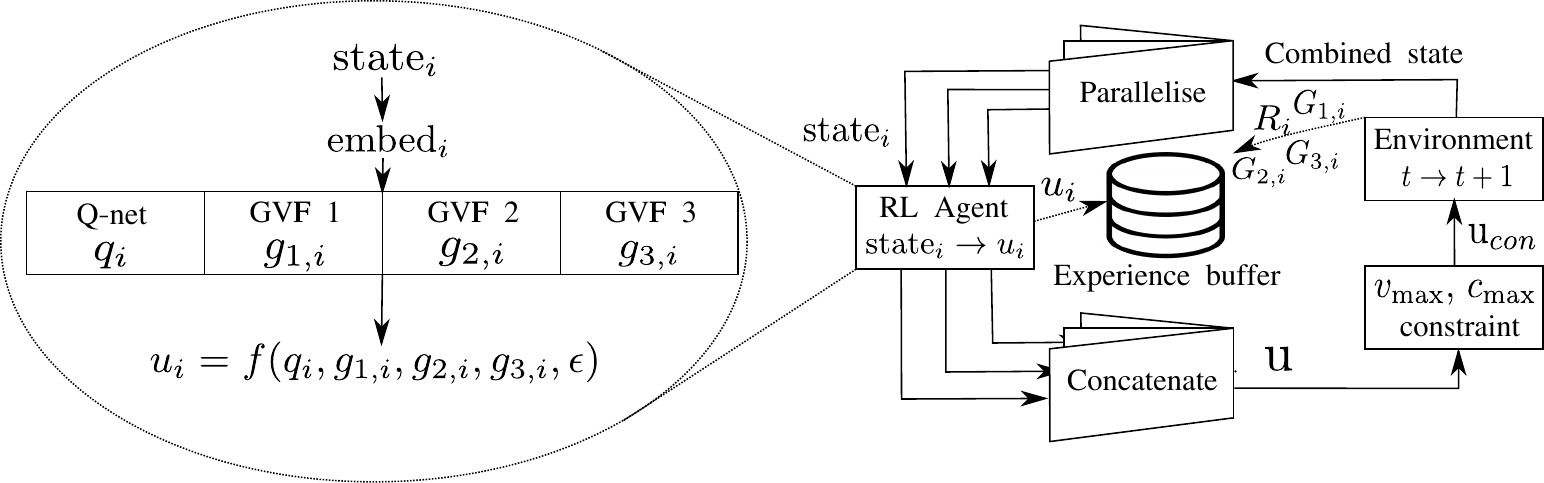}
    \caption{Implementation of DEZ-Greedy in the proposed problem. Copies of the RL agent process the individual $\mathrm{state}_i$ for each product $i$.}
    \label{fig:setup}
\end{figure}

We consider a retail store that offers a variety of $P$ products (Figure \ref{fig:setup}), and the inventory of each product is denoted as $x_{i}(t)$ at time step $t$ (where $1\leq i \leq P$), is closely monitored. The store has the flexibility to request an order in each time step, specifying a quantity, $u_{i}(t)$ to be replenished from the warehouse. Each product is subject to a maximum storage shelf limit, $M_{i}$. Both $x_{i}$ and $u_{i}$ are scaled variables, calculated by dividing the actual product quantity by $M_{i}$. We assume that the time gap between placing an order and receiving the replenishment is negligible compared to the duration of each time step. Given these consideration, the state is updated as follows:
\begin{equation}
    {\bf{x}}(t)^+ = {\bf{x}}(t)^- + {\bf{{u}}}(t) 
    \label{eq:repldyn}
\end{equation}
We have $P$-sized vectors denoted as ${\bf{x}}(t)^-$, ${\bf{x}}(t)^+$, and ${\bf{{u}}}(t)$, which pertain to all the products. These vectors are scaled based on the maximum inventory capacity for each respective product. The superscripts $+$ and $-$ signify the inventory level immediately prior to and following replenishment. The store's limitations are defined by the following constraints:  
\begin{align}
{\bf{0}} \leq {\bf{x}}(t) & \leq {\bf{1}}, \label{eq:inv} \\
{\bf{0}} \leq {\bf{u}}(t) & \leq {\bf{1}} - {\bf{x}}(t)^-,  \label{eq:control} \\
%{\bf{0}} \leq {\bf{x}}_j(t)^- + {\bf{u}}_j(t) & \leq {\bf{1}}, \label{eq:shelf} \\
{\bf{v}}^\intercal\,{\bf{u}}(t) \leq v_{\mathrm{max}} & \quad\text{ and }\quad
{\bf{c}}^\intercal\,{\bf{u}}(t) \leq c_{\mathrm{max}}.  \label{eq:truckvol}
\end{align}
In this context, constraint (\ref{eq:inv}) places a restriction on the normalized inventory level for each individual product. Constraint (\ref{eq:control}) specifies that the amount of products requested during any given moment must not surpass the remaining shelf capacity for those specific products at that time. Constraint (\ref{eq:truckvol}) establishes an upper limit, denoted as $v_{\mathrm{max}}$, on the aggregated volume, and $c_{\mathrm{max}}$ on the aggregated weight, considering all the products, respectively. This constraint effectively accounts for the limitations to transportation capacity. The relationship between inventory levels from on time step to the next is described as follows:
\begin{equation}
    {\bf{x}}(t+1)^- = \max \left( {\bf{0}}, {\bf{x}}(t)^+ - {\bf{w}}(t, t+1) \right),
    \label{eq:invchange}
\end{equation}
Here, ${\bf{w}}(t,t+1)$ represents the vector indicating the quantities requested for each product between time steps $t$ and $(t+1)$. The objective of the replenishment algorithm is to calculate the ideal replenishment quantity, denoted as ${\bf{{u}}}(t)$ which aims to maximize a 'business reward' based on the following explanation.

Inventory management poses a multi-objective optimization challenge, involving explicit costs associated with two primary aspects: (1) the risk of depleting inventory to zero, commonly referred to as "out-of-stock," and (2) the quantity $q_{waste,i}(t)$ of products that become wasted or spoiled during the time period ending at t. Additionally, we aim to ensure equitable treatment among products, even when the system faces stress, such as when capacities like $v_{\mathrm{max}}$ and $c_{\mathrm{max}}$ cannot adequately keep pace with product demand. To address this, we introduce a fairness penalty that considers the variation in inventory levels across all products, spanning from the 95th to the 5th percentile, denoted as $\Delta {\bf{x}}(t)^\mathrm{.95}_{\mathrm{.05}}$. Furthermore, we incorporate a penalty term $\Omega(t)$, which encompasses the total refused orders during the specified time interval, aggregating across all products. The objective, which we seek to maximize, is defined in equation (\ref{eq:business_rwd}). Given that no control interventions are allowed between consecutive time steps, all terms can be considered as aggregate values received at time $t$. The objective function is expressed as follows:
\begin{equation}
    \begin{split}
        R(t) = 1 - \underbrace{\frac{p_\mathrm{empty}(t)}{p}}_{\text{Out of stock}} & - \underbrace{\frac{p_\mathrm{critical}(t)}{p}}_{\text{Critical level}} - \underbrace{\frac{\sum_{i}q_\mathrm{waste,i}(t)}{p}}_{\text{Wastage}} \\ & - \underbrace{\Delta {\bf{x}}(t)^\mathrm{.95}_{\mathrm{.05}}}_{\text{Percentile spread}} - \underbrace{\Omega(t)}_{\text{Refused orders}}
    \end{split}
    \label{eq:business_rwd}
\end{equation}
where $p$ represents the total count of products (corresponding to the size of $\bf{x}$, $p_\mathrm{empty}(t)$ signifies the quantity of products with $x_{i} = 0$ at the conclusion of time interval $[t - 1, t)$. $p_\mathrm{critical}(t)$ is the number of products with $x_{i} < \kappa_{i}$ at the end of the same time interval, ensuring that each product maintains at least its minimum presentation level $\kappa_{i}$ or a critical inventory level. Since the highest possible value for $p_\mathrm{empty}(t)$, $p_\mathrm{critical}(t)$ and $\Delta(t)$ equals $p$, the maximum potential value for $q_\mathrm{waste,i}(t - 1, t)$ is 1. Consequently, the theoretical range for the objective/reward lies within the interval of -4 to 1. For practical purposes, individual terms are expected to be smaller than 1, and the majority of the reward should fall in the range of -1 to 1. The primary aim of the algorithm is to optimize the discounted cumulative sum of this reward defined in equation (\ref{eq:business_rwd}), while adhering to the constraints detailed in equations (\ref{eq:inv}) to (\ref{eq:truckvol}). 

% where ${\bf{w}}(t,t+1)$ is the vector of the quantities demanded for each product between $t$ and $(t+1)$. The goal of the replenishment algorithm is to compute the optimal replenishment quantity ${\bf{{u}}}(t)$ that maximises a `business reward' as per the following description.

\section{Methodology}

Our solution makes use of DQN \citep{mnih2015human} which is an popular off-policy algorithm where we use the previous policy samples to improve it's value estimation. Along with DQN we define GVFs to estimate a domain-critical property and use them as sub-policies to learn predictive knowledge of the system. We also utilize the Directed EZ-greedy strategy \citep{kalwar2022follow}, as another baseline. We compare these different algorithms for our study and show their performance across different number of products. 

We note that the order rate $\bf{w}(z)$ plays a key role in the system dynamics (\ref{eq:truckvol}). For simplicity, we define an estimator for the sales rate $w_i$ of each product $i$ in the form of a trailing average of sales in the most recent $T$ time periods, 
%
%%%%%%%%% Enter forecast equation 
\begin{equation}
    \hat{w_{i}}(z) = \frac{\int_{t-T}^{t} w_{i}(z^\prime) \,dz^\prime}{T}, \forall z \in [t, t + 1) i \in {1, ...,p}
    \label{eq:forecast}
\end{equation}
It must be remarked that the above forecasting algorithm returns a very crude approximation of the actual demand. Instead, one can leverage more sophisticated forecasting algorithms \citep{barlas2011demand, carbonneau2008application, hofmann2018big}, such as recurrent neural network(RNN)-based approaches to provide a better estimate of the true demand; however, design of the most appropriate forecasting algorithm is not the main focus of this paper. Instead, we rely on the simpler averaging based forecasting algorithm and still achieve near-optimal performance for inventory control. A better forecasting algorithm would only aid the RL algorithm in further improving its capability towards optimal replenishment decision. Since we assume that each period lasts for a unit of time, the forecast for aggregate orders, $W_{i}(t)$, is also given by (\ref{eq:forecast}).

The primary challenges in applying RL to this problem are (a) the large number of products $p$, (b) handling the shared capacity constraints (\ref{eq:truckvol}), and (c) the fact that the number of products can change over time. We describe an algorithm for parallelised computation of replenishment decisions, by cloning the parameters of the same RL agent for each product and computing each element of the vector $\bf{u}(t)$ independently. The advantage of this approach is that it splits the original problem into constant-scale sub-problems. Therefore, the same algorithm can be applied to instances where there are a very large (or variable) number of products. Despite parallelisation, we handle the shared capacity constraints as follows.

\subsection{Rewards}
The computation of individual components of $\bf{u}$ faces two primary difficulties: firstly, guaranteeing compliance with the system-level constraints outlined in equation (\ref{eq:truckvol}), and secondly, ensuring equitable treatment of all products. These challenges are mitigated to some extent through the reward system. The fairness concern is specifically addressed by incorporating the percentile spread term mentioned in equation (\ref{eq:business_rwd}), as it penalizes the agent when certain products have low inventory levels while others are maintained at higher levels. Moreover, the constraints related to volume and weight are introduced as flexible penalties in the subsequent definition of the 'per-product' reward, which is customized for individual decision-making purposes. 
%
%%%%% Enter NN Reward equation here
\begin{equation}
    \begin{split}
        R_i(t) = 1 - b_{i,empty}(t) & - b_{i,critical}(t) - q_{waste,i}(t) \\ - \Delta{\bf{x}}(t)^\mathrm{.95}_{\mathrm{.05}} 
        & - \Omega_i(t) - \alpha \max(\rho - 1, 0)
    \end{split}
    \label{eq:nn_rwd}
\end{equation}
where $b_{i,empty}(t)$ and $b_{i,critical}(t)$ are binary variables that serve as indicators, determining whether the inventory for product $i$ falls below 0 or $\kappa_i$ in the current time period. Additionally, $\alpha$ stands as a fixed parameter, while $\rho$ represents the ratio of the total volume or weight requested by the RL agent to the existing capacity. This can be precisely defined as follows:
%
%%%%% Enter rho equation 
\begin{equation*}
    \rho = \max \left( \frac{\bf{v^{\intercal}u}(t)}{v_{max}}, \frac{\bf{c^{\intercal}u}(t)}{c_{max}} \right)
\end{equation*}
Equation (\ref{eq:nn_rwd}) outlines the reward provided to the RL agent, which differs from the genuine system reward described in (\ref{eq:business_rwd}). When the combined actions generated by the agent for all products do not surpass the available capacity ($\rho \leq 1)$, then the average value of (\ref{eq:nn_rwd}) equals that of (\ref{eq:business_rwd}). This suggests that maximizing $R_i(t)$ is essentially the same as maximizing $R(t)$, as long as the system constraints remain unbroken. The final two components of (\ref{eq:nn_rwd}) are consistent for all products at the given time $t$.

\begin{table}
    \centering
    \begin{tabular}{l|l}
        \toprule
        Notation                            & Description \\
        \midrule
        $x_i(t)$                            & Current inventory level \\ 
        $\hat{W_i}(t)$                      & Forecast aggregate orders in $[t, t + 1)$ \\
        $v_i$                               & Unit volume \\
        $c_i$                               & Unit weight \\
        $T_s(i)$                            & Shelf-life \\
        $\bf{v^{\intercal}\hat{W}(t)}$      & Total volume of forecast for all products \\ 
        $\bf{c^{\intercal}\hat{W}(t)}$      & Total weight of forecast for all products \\      
        \bottomrule
    \end{tabular}
    \caption{State space representation}
    \label{tab:state_space}
\end{table}

\subsection{State and action space}
Table \ref{tab:state_space} presents a list of features utilized to calculate the replenishment quantity for each product. The initial two features pertain to the current state of the system concerning product $i$. The subsequent three inputs contain product-related information, encompassing both long-term and unchanging behavior. The parameter shelf-life $T_s(i)$, is a normalized measure inversely proportional to the average inventory reduction for product $i$ signifying the decrease in inventory not explained by orders during a specified time frame. 
% This value is determined empirically and serves as an implicit estimator for the dynamics represented by $A$. 
These metadata elements are used to distinguish between different product characteristics when they are processed sequentially by the same RL agent, achieved by mapping individual products into a common feature space. The final two features listed in Table \ref{tab:state_space} are derived indicators that offer insights into the overall demand on the system, considering various constraints. These indicators function as constraints on the control action for product $i$ when the system experiences high demand. They also aid the agent in establishing a connection between the last term in the observed rewards (the penalty for exceeding capacity) and the input states. The outcome of the RL agent is $u_i(t)$ which represents the intended action for product $i$ at time $t$ Individual actions are combined to form $\bf{u}(t)$ as depicted in the workflow illustrated in Figure \ref{fig:im_problem}. 

\subsection{General Value Functions}
The general value function is a more generalized definition of the value function in reinforcement learning. It can be learned for any predictive signal not just for reward signal. In literature, these predictive signals are also known as cumulants $C$. These cumulants can be thought of as queries directed towards GVFs. Formally, GVFs are defined as the expected cumulative discounted sum of cumulants for a given policy $\pi$ and a state-action pair $(s, a)$: \begin{equation} 
Q^{GVF}(s_t,a_t) = \mathbb{E}_{\pi} [\sum_{t=0}^{\infty} \gamma^{t}C_t(s_t,a_t)]
\end{equation}
Where $\gamma$ is a discount factor. And GVFs can be learned by any value-based RL algorithm e.g. temporal difference (TD) learning.\\ 
For this environment setup we used 3 GVFs which are trained on following cumulant signals defined as -
\begin{equation}
	C_{1,i}(t) = q_{waste,i}(t)
\end{equation} 
where, $q_{waste,i}(t)$ is the wastage quantity of product $i$ during the time period end at $t$. The GVF learned on cumulant signal $C_{1,i}(t)$, we denote it by $Q^{GVF_1}$ and its policy by $\pi^{GVF_1} = \underset{a}{\mathrm{argmin}}\,Q^{GVF_1}(s,a)$.
\begin{equation}
	C_{2,i}(t) = \begin{cases}
		1, & \text{if $x_i(t)=0$}.\\
		0, & \text{otherwise}.
	\end{cases}
\end{equation} 
Cumulant $C_{2,i}(t)$ represents the product $i$ going out of stock at the end of time-period $[t-1,t)$. The GVF learned on cumulant signal $C_{2,i}(t)$, we denote it by $Q^{GVF_2}$ and its policy by $\pi^{GVF_2}=\underset{a}{\mathrm{argmin}}\,Q^{GVF_2}(s,a)$.
\begin{equation}
	C_{3,i}(t) = 1 - x_i(t) 
\end{equation} 
Cumulant $C_{3,i}(t)$ encodes the depletion in inventory of product $i$ during the time period end at $t$. The GVF learned on cumulant signal $C_{3,i}(t)$, we denote it by $Q^{GVF_3}$ and its policy by $\pi^{GVF_3}=\underset{a}{\mathrm{argmin}}\,Q^{GVF_3}(s,a)$.

\begin{figure*}
    \centering
    \includegraphics[width=.9\linewidth]{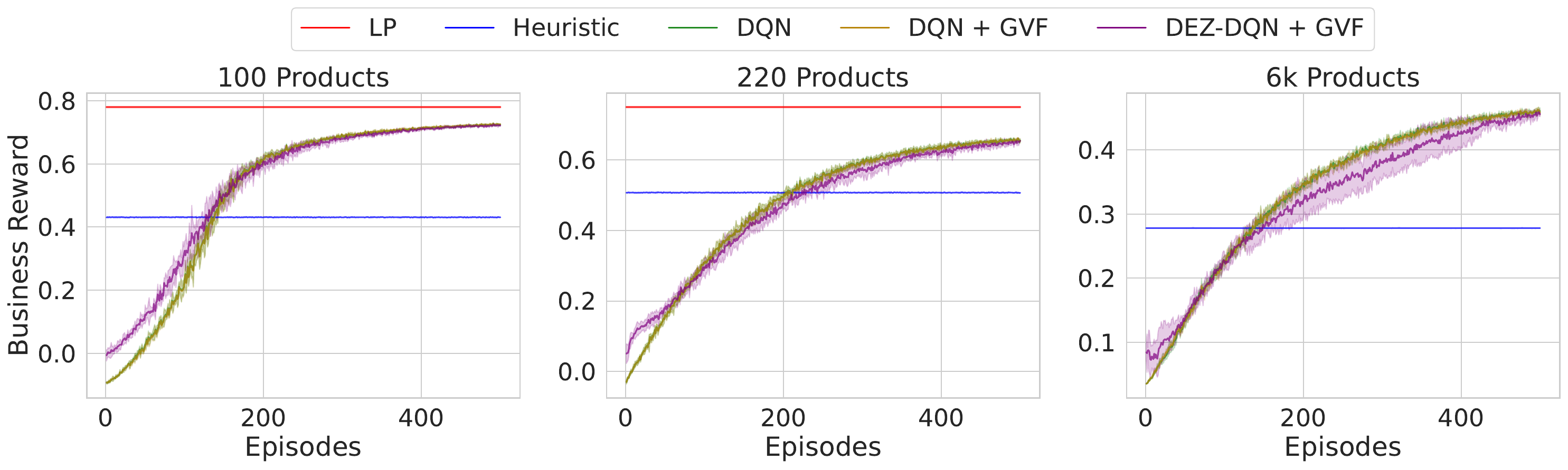}
    \caption{Comparing algorithm training performance on varying dataset sizes: Mean results across 5 random seeds with 95\% confidence interval shading for $100$, $220$, and $600$ product datasets}
    % \caption{Training performance over business reward of different algorithms for $100$, $220$, and $6k$ product datasets. Solid line represents mean across $5$ random seeds and shaded region denotes $95\%$ confidence interval.}
    \label{fig:100_prod BR}
\end{figure*}

\subsection{Directed EZ-Greedy Strategy}
Directed EZ Greedy utilizes General Value Functions (GVFs) as auxiliary tasks and employs sub-policies derived through greedy action selection for a temporally extended $\epsilon$-greedy approach. It occasionally chooses a random option (with probability $\epsilon$), possibly originating from any of the randomly selected GVFs in the current state. It then continues to take the greedy action based on that GVF, facilitating directed exploration across various state regions. This distinctive exploration strategy is why it's called Directed EZ Greedy. Algorithm \ref{alg:DEZ} is adapted from \citep{kalwar2022follow} and in this paper we kept the maximum persistence to $1$.

\begin{algorithm}
\caption{DEZ-Greedy exploration strategy}
\SetKwFunction{FMain}{\text{DEZGreedy}}
\SetKwProg{Fn}{\text{Function}}{:}{}
%\Fn{\FMain{$\epsilon$, $Z_{max}$}}{
\Fn{\FMain{$\epsilon$}}{
%Countdown timer $z \leftarrow 0$ \\
%Uniformly sampled random action $w \leftarrow -1$ \\ 
Selected GVF index $g \leftarrow 0$ \\
\While{\text{True}}{
    \text{Observe state}  $s$ \\
    %\eIf{$z==0$}
    %{
        % // Timer ran out: choose controlling policy\\
        \eIf{$random() < \epsilon$}{ // Explore \\
        %\text{Sample duration:} $z \sim [1, Z_{max}]$ \\
        \text{Sample GVF:} $g \sim [0,M]$ \\
        \eIf{$g==0$}
        %{\text{Sample action:} $w \leftarrow U(A)$ \\ $a \leftarrow w$}
        {\text{Sample action:} $a \leftarrow U(A)$}
        {$a \leftarrow argmin(Q^{GVF}_g)$}
        }
        {// Exploit \\ $a \leftarrow argmax(Q^{Main})$}
    %}
    % {
    %     // Continue previous exploration policy \\
    %     \eIf{$g==0$}{$a \leftarrow w$}{$a \leftarrow argmax(Q^{GVF}_g)$}
    %     $z \leftarrow z-1$
    % }
    \text{Take action} $a$
    }
}
\label{alg:DEZ}
\end{algorithm}

%%%%%%%%%%%%%%%%%%%%%%%%%%%%%%%%%%%%%%%%%%%%%%%%%%%%%%%%%%%%%%%%%%%%%%%%%%%%%%%%%%%%%%%%%%%%%%%%%%%%%%%%%%%%%%%%%%%%%%%%%%%%%%%%%

\section{Experimental details}
\subsection{Data for experiments}
As our methodology aligns with \citep{meisheri2022scalable}, we use a similar procedure to generate out data. \citep{meisheri2022scalable} conducted experiments on publicly available data set for brick and mortar stores \citep{Kaggle}. The original data set encompasses purchase information for 50,000 different product types and involves 60,000 unique customers. However, it lacks metadata about products such as dimensions and weight. To align this data with the format required for their work, they attributed dimensions and weights to each product type based on the product descriptions available. This process was carried out manually, so \citep{meisheri2022scalable} only used two distinct subsets containing 220 and 100 products, respectively, from the complete data set. We formulated an algorithmic approach to generating and assigning metadata, allowing us to extend the experiments up to 6000 products and we did the experiments on all three datasets containing 6k, 220 and 100 products. These subsets are specifically chosen from products categorized as 'grocery' from original data set \cite{Kaggle}.

All three datasets covers a span of 349 days and we assume that stock replenishment occurs four times daily, resulting in a total of 1396 time periods. From these, we allocate the initial 900 time periods for training purposes and reserve the remaining 496 time periods for testing. We utilize three semi-synthetic data sets, each consisting of 6k, 220 and 100 product types, to compare against other approaches. We set the volume and weight constraints (denoted as vmax and cmax) slightly below the average sales volume and weight to ensure that constraints \ref{eq:truckvol} remain active.

For RL based algorithms, we discretized  the replenishment values for each product $i$ in $14$ discrete actions (0, 0.005, 0.01, 0.0125, 0.015, 0.0175, 0.02, 0.03, 0.04, 0.08, 0.12, 0.2, 0.5, 1). These actions represent the normalized replenishment quantities relative to the maximum shelf capacity for each respective product. During each training episode, we generate random initial inventories for each product. These random initial inventories are used for all the experiments across algorithms (RL, heuristic, linear programming). We ran experiments for 5 random seeds for statistical significance.

\subsection{Baseline algorithms}

\subsubsection{\textbf{Heuristic based on proportional control:}} 
We employ a customized version of the s-policy \cite{nahmias1994optimizing}, a well-established heuristic in the literature aimed at maintaining inventories at predefined constant levels. In this heuristic, denoting $x^*$ as the target inventory level for products, the replenishment quantity is carefully designed to meet forecasted sales and bring the inventory to the desired level $x^*$ by the end of the time period. This is mathematically expressed as:
\begin{equation}
    u_{pr}(t) = \max[0, x^* + \hat{W} - x(t)^-]
    \label{eq:heu_action_selection}
\end{equation}
The action derived from Eq. (\ref{eq:heu_action_selection}), denoted as $u_{pr}(t)$, satisfies constraints (\ref{eq:inv}) and (\ref{eq:control}) since $0 \leq x^* \leq 1$. However, it is important to note that this approach does not guarantee compliance with the total volume and weight constraints specified in Eq. (\ref{eq:truckvol}).

\begin{figure*}
    \centering
    \includegraphics[width=\linewidth]{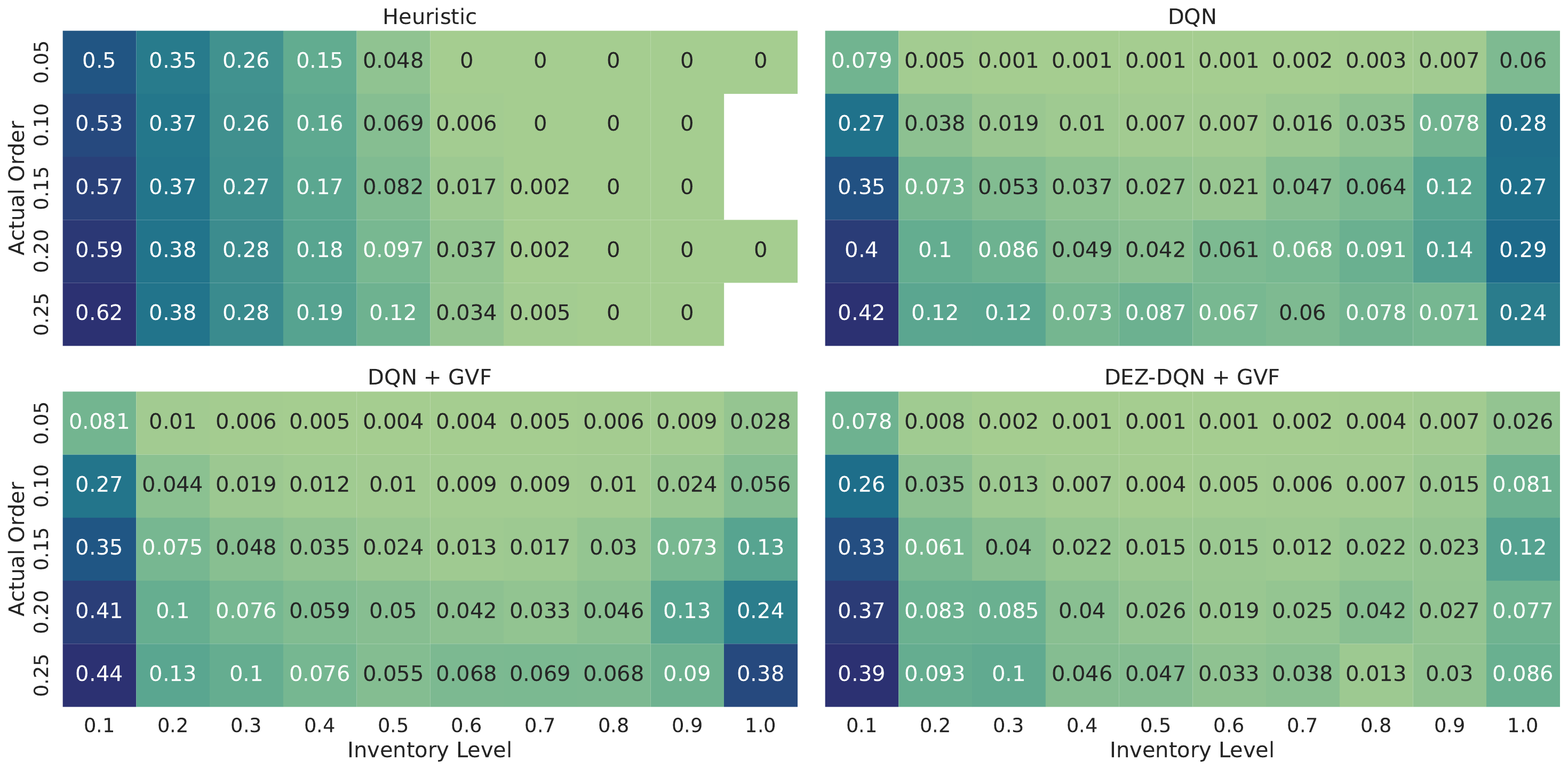}
    \caption{Visualizing main policy heatmaps: A comparative analysis across various algorithms using the 6k product dataset}
	% \caption{Comparison of all algorithm's main policy heatmap on testing samples of 6k product dataset.}
	\label{fig:6k_prod heatmap}
\end{figure*}

\subsubsection{\textbf{Optimal bounds using linear programming:}} 
%Linear programs (LP) and their robust variants have traditionally played a pivotal role in single-store, single-product inventory optimization [\cite{bertsimas2004robust}, \cite{bertsimas2006robust}, \citep{lee1996mixed}]. While LP approaches excel in such single-product scenarios, their applicability becomes limited in more complex multi-product constraints settings. Nonetheless, they can yield optimal performance bounds when implemented effectively. 
In this paper, we employ the same formulation as in \cite{meisheri2022scalable} to construct an online linear programming (LP) model with perfect information, where the true or realized demands $W$ are known beforehand. The LP's objective is to optimize a linear combination of components constituting the overall business reward defined in Eq. (\ref{eq:business_rwd}), while adhering to linear constraints outlined in Eqs. (\ref{eq:inv}) through (\ref{eq:truckvol}), along with a nonlinear constraint specified in Eq. (\ref{eq:inv}) that accounts for the total demand lost due to insufficient inventory. We linearize the maximum constraint in Eq. (\ref{eq:inv}) by introducing an additional LP variable. By leveraging perfect information to generate actions, the LP solution gives us with theoretical upper bounds for a given environment.

\section{Results and discussion}
In this section, we compare the performance of DEZ-DQN+GVF with $\epsilon$-greedy strategy as well as heuristic and LP based theoretical upper bound. Our analysis encompasses the training phase results for all algorithms, and we also evaluate and compare their performance on testing data. We evaluate all the algorithms over the business reward that is defined in the problem formulation.\\
\textbf{\textit{Algorithms Compared:}} In addition to the baseline algorithms discussed in the preceding section, we extend our analysis to include the following algorithms:\\
$\bullet$ \textit{DQN:} In our experimental setup, both the general value functions (GVFs) and the primary Q-value function are trained using the DQN framework \cite{mnih2015human}. Thus, our baseline for comparison is the DQN algorithm. DQN employs $\epsilon$-greedy exploration with a gradually annealing $\epsilon$ value, a strategy we also adopt for our experiments.\\
$\bullet$ \textit{DQN+GVF:} This algorithm leverages GVFs solely for representation learning through a shared representation, which is simultaneously updated by the primary DQN agent and all the GVFs. Similar to DQN, it incorporates $\epsilon$-greedy exploration for action selection.\\
$\bullet$ \textit{DEZ-DQN+GVF:} In this variant, GVFs not only contribute to improved representation learning but also play a role in action sampling for directed exploration. For a comprehensive understanding of the hyper-parameters used in all these algorithms, please refer to Appendix 1 in the supplementary materials.

\subsection{Training results}
From Figure \ref{fig:100_prod BR}, it is evident that the converged value for the averaged business reward is the same across DQN, DQN+GVF, and DEZ-DQN+GVF on 100, 220, and $6k$ products datasets. But they significantly outperform the proportional control-based heuristic and we also observe that they are able to achieve $92\%$, and $86\%$ of the LP-based upper bound for the 100, and 220 products datasets, respectively. The LP solver did not complete its optimization process within the allocated computational time for the $6k$ products dataset. Additionally, the dual gap is very large at the end of the computational period so LP Solver is not able to find feasible solution for $6$k product dataset. Here note that the LP-based upper bound provides us with theoretical upper bounds for a given environment as it uses the perfect information to generate actions. We can also observe that the DEZ-DQN+GVF have better-averaged business reward during the initial exploration  and achieve $70\%$(approx) of converged value faster compared to DQN and DQN+GVF for $100$ product dataset. Note that the DQN line is not clearly visible due to overlap with DQN+GVF. For a comprehensive understanding of the performance of all the algorithms, we provide a detailed component-wise analysis of the business reward in Appendix 2.
\subsection{Testing Results}
In Figure \ref{fig:6k_prod heatmap}, the heatmap depicts the mean replenishment quantity associated with specific inventory and order bins, where, on the x-axis, an inventory level of 0.2 denotes inventory levels ranging from 0.1 to 0.2, and on the y-axis, an order value of 0.10 represents order values between 0.05 and 0.10. The heatmap values are further averaged across five random seeds. (Please note: white space in the heatmap indicate values not available).

\begin{figure*}
    \centering
    \includegraphics[width=\linewidth]{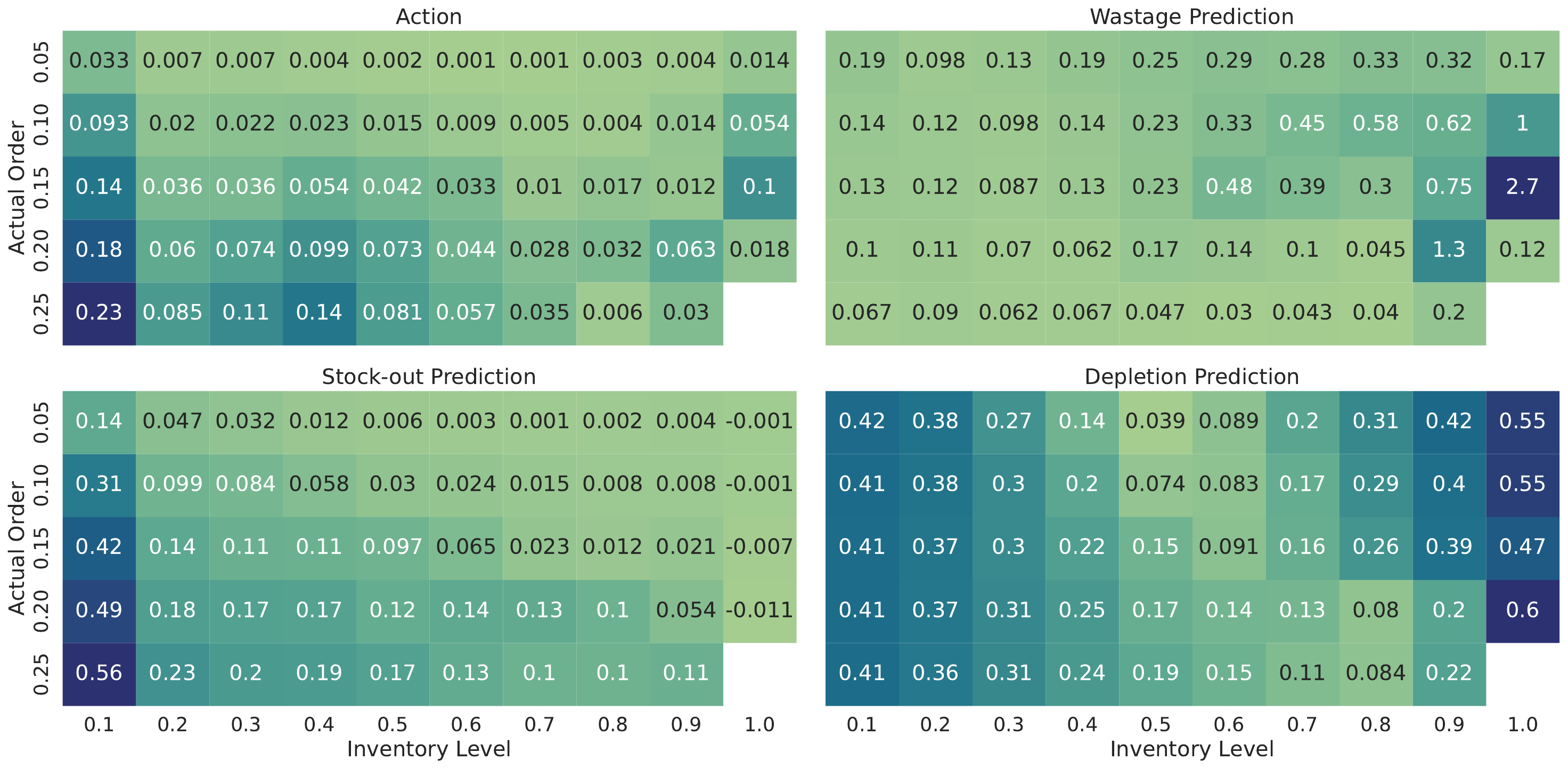}
    \caption{Visualizing averaged replenishment decisions and corresponding GVF predictions for the 220-product dataset: The top-left figure displays a heatmap of the averaged replenishment actions selected by DEZ-DQN+GVF's primary policy, while the other three figures show heatmaps of the averaged GVF quantity predictions (wastage, stock-out and depletion) corresponding to the chosen actions.}
	% \caption{Heatmap of averaged replenishment action chosen by DEZ-DQN+GVF's main policy on 220 products dataset (top left figure) and heatmaps of averaged prediction of GVF quantities corresponding to the chosen action from main policy of DEZ-DQN+GVF (other three figures).}
	\label{fig:220_prod heatmap}
\end{figure*}

Analysis of the heatmaps reveals a clear pattern: when the inventory level remains constant and order values increase, the replenishment quantity also rises. Conversely, when the order value remains fixed and inventory levels are higher, the replenishment quantity tends to be lower compared to situations with lower inventory levels. This consistent trend is observed across all the algorithms. It is worth noting that in all algorithmic heatmaps, there are instances on the right side where higher replenishment appears to occur despite high inventory levels. This happens mostly to due to fact that there is a higher variance in demand in the training data and we are using forecast demand in the state and not the actual demand of the products, so there is chance of going out of stocks if the demand spike is high enough.

Furthermore, the heatmap clearly shows that the DEZ-DQN+GVF agent consistently takes small replenishment actions compared to other algorithms. This behavior can be explained by the fact that DEZ-DQN+GVF explicitly learns the GVF corresponding to the wastage signal and strives to maintain low inventory levels because wastage is directly proportional to inventory levels. This distinction is not observed in the DQN+GVF heatmap, even though it also learns GVFs explicitly from the wastage signal. This disparity may be attributed to the difference in how GVFs are learned in the two algorithms. In DQN+GVF, GVFs are learned from off-policy data (trained on sampled actions from the main policy) and lack corrective feedback from the GVF's policy. In contrast, DEZ-DQN+GVF benefits from GVFs' policies for action sampling during exploration, which results in more efficient control over inventory levels. Furthermore, the heatmap plots corresponding to datasets containing 100 and 220 products can be found in Appendix 3 within the supplementary materials.

\subsection{Transfer learning, explainability, and generalisation}

\textbf{\textit{Transfer Learning:}} Table \ref{table:t&t perf} presents the training and testing performance of all algorithms over business reward averaged over 5 random seeds on all datasets. In order to assess the transfer capability of the learnt GVFs to new scenarios, we test all the algorithms on out-of-distribution datasets (ones they are not trained on). It is noteworthy that the datasets encompassing 100 products, 220 products, and 6k products are completely distinct, exhibiting no overlap between them. Our results demonstrate that DQN, DQN+GVF, and DEZ-DQN+GVF consistently outperform the heuristic even when trained on different datasets.

\textbf{\textit{Explainability:}} Additionally, we can gain insights into the actions chosen by the DEZ-DQN+GVF policy by examining the heatmaps of the GVF predictions. The three learned GVFs play a crucial role in illuminating essential system characteristics, offering predictive cues for selecting appropriate actions. This visualization is presented in the figure \ref{fig:220_prod heatmap}, which showcases the average replenishment decisions for the dataset of 220 products alongside the predictions of three GVFs (namely wastage, stock-out, and depletion) corresponding to inventory levels and actual demand. 

\begin{itemize}
    \item \textit{Wastage Prediction Heatmap:} Our model links wastage directly to inventory levels and product shelf-life. Consequently, the heatmap reveals that when inventory levels are low, wastage predictions are low, and vice versa. Additionally, when demand is high at a particular inventory level, wastage predictions tend to be low as well.
    \item \textit{Stock-out Prediction Heatmap:} Clearly, lower inventory levels coupled with high demand yield high stock-out predictions. As inventory levels increase, we observe a smooth transition in stock-out predictions, as the likelihood of running out of stock diminishes.
    \item \textit{Depletion Prediction Heatmap:} In this case, we aim to understand inventory depletion levels as we make replenishment decisions to maintain desired inventory levels. The heatmap indicates that lower inventory levels result in higher depletion predictions. Interestingly, there are instances where high depletion predictions occur even with high inventory levels. This phenomenon is attributed to the significant demand variability in the training data, which can lead to stock-outs if demand spikes are substantial.
\end{itemize}

Because these GVF predictions directly relate to tangible system attributes, they offer a clear and interpretable perspective for human supervisors. Consequently, the values of chosen actions from the primary policy can be explained using these GVF predictions.

\begin{table}
	   \centering
	   \begin{tabular}{|l|l||l|l|l|}
         \hline
		     Algorithm & Training & \multicolumn{3}{c|}{Testing}\\
		     \hline
		      &  & Self & \multicolumn{2}{c|}{Transfer learning}\\
           \hline
           \multirow{2}{*}{\textbf{100 products}} & & & Trained on & Trained on \\
           & &  &  220 prods & 6k prods\\
           \hline
          Heuristic   & 0.431 & 0.506 & -     & - \\
		     DQN         & 0.722 & 0.723 & 0.708 & 0.708 \\
		   	 DQN+GVF     & 0.723 & 0.727 & 0.716 & 0.707 \\
		   	 DEZ-DQN+GVF & 0.720 & 0.724 & 0.717 & 0.709 \\
          LP-upper bound & 0.780 & 0.780 & - & -\\
		   \hline
		   	 \multirow{2}{*}{\textbf{220 products}} & & & Trained on & Trained on\\
           & &  & 100 prods & 6k prods\\
           \hline
          Heuristic   & 0.507 & 0.493 & -     & - \\
		     DQN         & 0.653 & 0.649 & 0.614 & 0.636 \\
		   	 DQN+GVF     & 0.653 & 0.654 & 0.602 & 0.626 \\
		   	 DEZ-DQN+GVF & 0.646 & 0.651 & 0.620 & 0.632 \\
          LP-upper bound & 0.749 & 0.749 & - & -\\
          \hline
		   	 \multirow{2}{*}{\textbf{6k products}} & & & Trained on & Trained on\\
           & &  & 100 prods & 220 prods \\
           \hline
          Heuristic   & 0.345 & 0.345 & -     & - \\
		     DQN         & 0.457 & 0.573 & 0.519 & 0.533 \\
		   	 DQN+GVF     & 0.457 & 0.536 & 0.534 & 0.541 \\
		   	 DEZ-DQN+GVF & 0.450 & 0.567 & 0.535 & 0.554 \\
          LP-upper bound & DNF & DNF & - & -\\
          \hline
		   \end{tabular}
     \caption{Training and testing performance over business reward averaged over $5$ random seeds for all datasets. Note that training and testing results are on 900 and 496 samples respectively, leading to higher average rewards on testing data sets. While the three RL based approaches have similar saturation rewards, their rates of convergence are different.}
	   \label{table:t&t perf}
\end{table}

\begin{figure}
    \centering
	\includegraphics[width=0.98\linewidth]{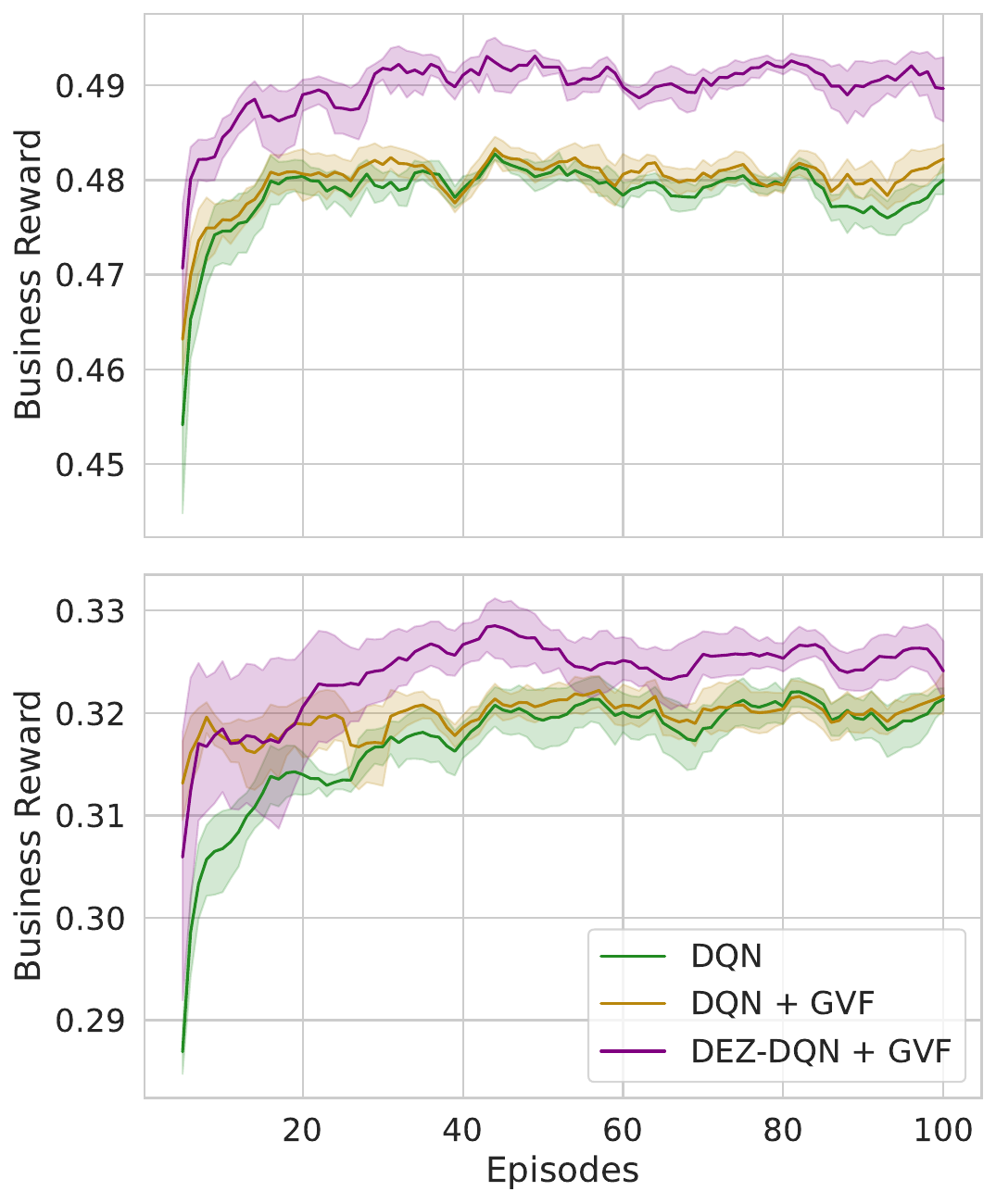}
	\caption{Generalisation on the 100-product dataset with the modified business reward definitions [Note: upper figure - wastage component of the business reward is changed, lower figure - empty level component of the business reward is changed.]. The solid line depicts the mean performance across $5$ random seeds, while the shaded region represents a $95\%$ confidence interval.}
	\label{fig:100_prod TF BR}
\end{figure} 

\textbf{\textit{Generalisability:}} Finally, the generalisability of the GVF approach can be demonstrated by the ability of the algorithm to adapt to significant deviations in the reward structure. Figure \ref{fig:100_prod TF BR} shows the training performance on the 100 product dataset for modified reward definitions. In the figure on the left, agents are fine-tuned on a modified reward in which the weightage on wastage is increased by a factor of 4. In the figure on the right, agents are fine-tuned on the modified reward in which the out-of-stock penalty is given at an inventory level of $0.1$ instead of the original $0.05$. We do the fine-tuning for 100 episodes with a constant exploration of $0.1$, starting from the original trained models. From the figure, it is evident the DEZ-DQN+GVF has a better performance compared to DQN and DQN+GVF, which shows that the DEZ-DQN+GVF can adapt to different business reward definitions. For a comprehensive understanding of the performance of three algorithms, we provide a detailed component-wise analysis of the business reward in Appendix 4. 

\section{Conclusion and future work}
In conclusion, we showed that the use of GVFs can be extended to realistic problems such as inventory management. Using the GVFs gave us better exploration leading to faster convergence, although the saturation reward levels in training were similar. However, the GVFs also gave us the ability to adapt to new tasks within the same environment, such as an abrupt change in the reward structure.
In our current work we have hand-designed attributes fed as cumulants for learning the system dynamics, in future we plan to instinctively identify the crucial attributes from the environment and use these cumulants for much better predictive decisions.   
% %%%%%%%%%%%%%%%%%%%%%%%%%%%%%%%%%%%%%%%%%%%%%%%%%%%%%%%%%%%%%%%%%%%%%%%%%%%%%%%%%%%%%%%%%%%%%%%%%%%%%%%%%%%%%%%%%%%%%%%%%%%
%%%%%%%%%%%%%%%%%%%%%%%%%%%%%%%%%%%%%%%%%%%%%%%%%%%%%%%%%%%%%%%%%%%%%%%%%%%%%%%%%%%%%%%%%%%%%%%%%%%%%%%%%%%%%%%%%%%%%%%%%%%

% %%% The next two lines define, first, the bibliography style to be 
% %%% applied, and, second, the bibliography file to be used.

\bibliographystyle{ACM-Reference-Format} 
\bibliography{sample}

%%%%%%%%%%%%%%%%%%%%%%%%%%%%%%%%%%%%%%%%%%%%%%%%%%%%%%%%%%%%%%%%%%%%%%%%

\end{document}